\documentclass{article}
\usepackage{ijcai22}

\usepackage{times}
\usepackage{soul}
\usepackage{url}
\usepackage[hidelinks]{hyperref}
\usepackage[utf8]{inputenc}
\usepackage[small]{caption}
\usepackage{graphicx}
\usepackage{amsmath}
\usepackage{amsthm}
\usepackage{booktabs}
\usepackage{algorithm}
\usepackage{algorithmic}
\usepackage{algorithm}
\usepackage{algorithmic}
\usepackage{amssymb}
\usepackage[switch]{lineno}
\usepackage{marvosym}

\title{Learning Coated Adversarial Camouflages for Object Detectors}

\author{
Yexin Duan$^{1,2}$\and
Jialin Chen$^3$\and
Xingyu Zhou$^4$\and
Junhua Zou$^2$\and
Zhengyun He$^{2,5}$\and
Jin Zhang$^1$\and
Wu Zhang$^2$\And
Zhisong Pan$^2$\footnote{Corresponding author}
\affiliations
$^1$Department of Watercraft Power, Army Military Transportation University of PLA, Zhenjiang, China\\
$^2$College of Command and Control Engineering, Army Engineering University
of PLA, Nanjing, China\\
$^3$The 28th Research Institute of China Electronics Technology Group Corporation, Nanjing, China \\
$^4$Communication Engineering College, Army Engineering University of PLA, Nanjing, China\\
$^5$Railway Transportation College, Hunan University of Technology, Zhuzhou, China
\emails
duanyexin0713@163.com,
hotpzs@hotmail.com
}

\begin{document}

\maketitle

\begin{abstract}
An adversary can fool deep neural network object detectors by generating adversarial noises. Most of the existing works focus on learning local visible noises in an adversarial ``patch" fashion. However, the 2D patch attached to a 3D object tends to suffer from an inevitable reduction in attack performance as the viewpoint changes. To remedy this issue, this work proposes the {\bf Coated} Adversarial Camouflage (CAC) to attack the detectors {\bf in arbitrary viewpoints}. Unlike the patch trained in the 2D space, our camouflage generated by a conceptually different training framework consists of 3D rendering and dense proposals attack. Specifically, we make the camouflage perform 3D spatial transformations according to the pose changes of the object. Based on the multi-view rendering results, the top-{\it n} proposals of the region proposal network are fixed, and all the classifications in the fixed dense proposals are attacked simultaneously to output errors. In addition, we build a virtual 3D scene to fairly and reproducibly evaluate different attacks. Extensive experiments demonstrate the superiority of CAC over the existing attacks, and it shows impressive performance both in the virtual scene and the real world. This poses a potential threat to the security-critical computer vision systems.

\end{abstract}

\section{Introduction}

Despite deep neural networks have achieved remarkable performance on various visual recognition tasks~\cite{szegedy2016rethinking,redmon2018yolov3}, they are found to be vulnerable to adversarial examples \cite{szegedy2014intriguing}, inputs with adversarial noises, which can fool the deep models without impeding human recognition. The adversarial attacks pose serious concerns in security-critical areas, such as medical diagnosis \cite{zhou2019collaborative} and autonomous driving \cite{sitawarin2018darts}.


\begin{figure}[t]
\centering
\includegraphics[width=0.95\columnwidth]{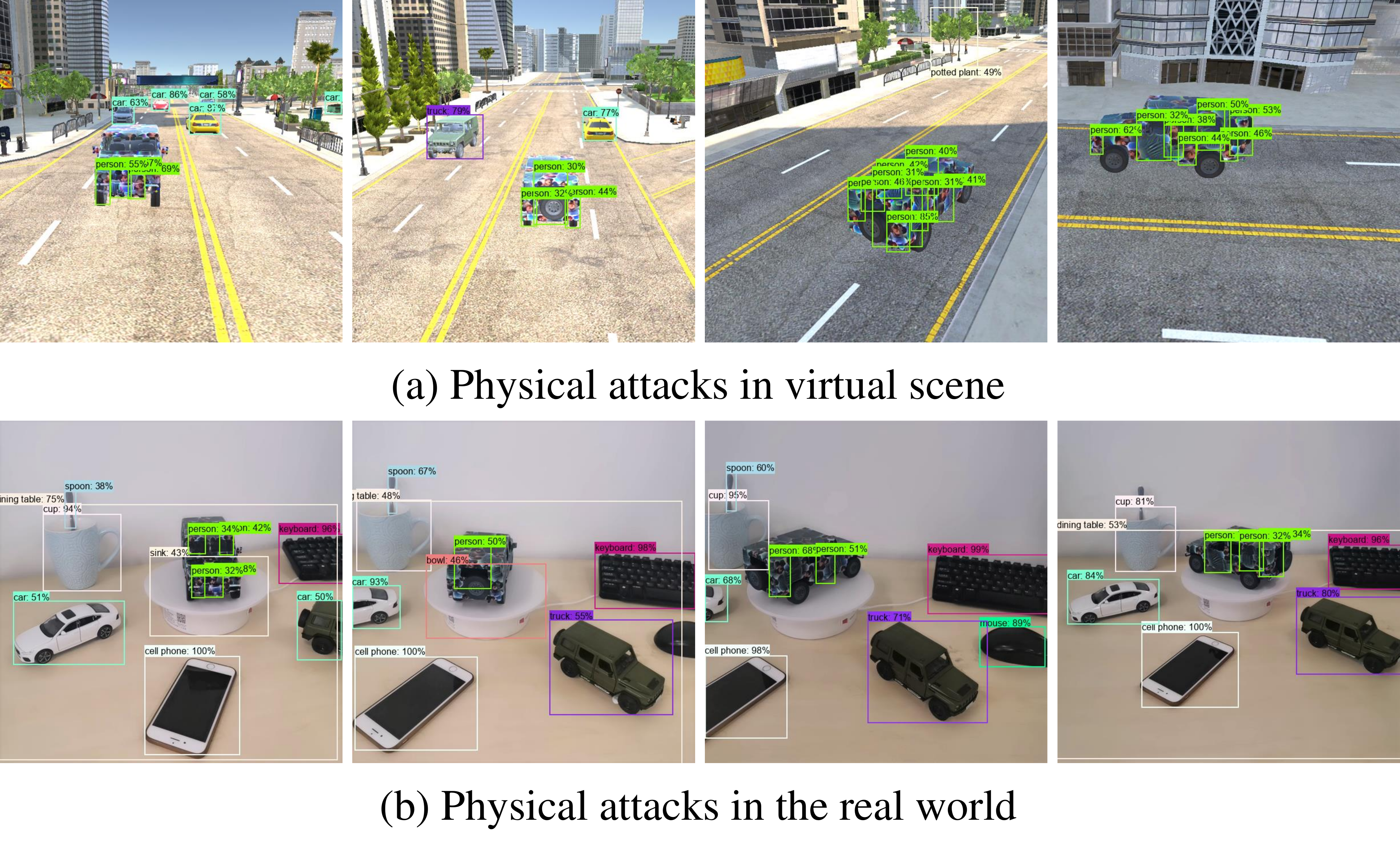} 
\caption{Physical attacks (CAC) against the Faster R-CNN detector in (a) 3D virtual scene and (b) real world in free viewpoints under different brightness conditions. The adversarial vehicles are detected as the target label ({\it e.g.}, person). Zoom in for more details.}
\label{fig:1}
\end{figure}

Attacks can be classified by the type of outcome the adversary desires: (1) non-targeted attack, the adversary's goal is to cause the deep model to predict any incorrect label; (2) targeted attack, the adversary aims to change the deep model's prediction to some specific target class, which is more challenging. In application domains, the attacks can be divided into digital attacks and physical attacks, with the latter posing a greater threat to real-world systems \cite{brown2017adversarial}. Compared with previous works which focus on generating adversarial objects for image classifiers \cite{athalye2018synthesizing}, attacking object detectors is a more realistic computer vision scenario, and is significantly harder as it requires misleading the classification results in multiple bounding boxes with different scales \cite{huang2020universal}.

Most of the existing works generate local visible adversarial patches to conduct physical attacks for object detectors. However, there are several limitations: (1) the adversarial patches are trained in two-dimensional space \cite{thys2019fooling,huang2020universal}, that is, during the training process, they do not carry out corresponding spatial transformations according to the pose changes of the objects, but only perform 2D plane transformations on the training images, resulting in the significant decline of the attack effectiveness when the viewpoint changes in the 3D physical world; (2) the adversarial noises are only for the planar objects, such as stop sign \cite{song2018physical,chen2018shapeshifter}, and they also would be less effective for an object with arbitrary view angles; (3) misidentifying the adversarial object as any incorrect class, not as a specific target class \cite{zhang2019camou} ({\it e.g.}, for autonomous driving, misidentifying an object as a designated person or stop sign is more threatening than randomly misidentifying it as a cake); (4) the missing of a unified physical evaluation environment makes it difficult to fairly evaluate the results of different attacks \cite{huang2020universal}. These limitations make attacks in the 3D physical world less effective and difficult to evaluate accurately.

To address these issues, we propose the Coated Adversarial Camouflage (CAC) attack, which generates an adversarial camouflage that covers the entire object. A combination of spatial transformations is utilized to render various poses of the object as well as lighting and other natural variations to eliminate the adversarial blind spots, and enable the adversarial camouflage to mislead the detector to recognize the object as a specific target class from any viewpoint in different environments. In addition, inspired by the diverse input strategy, which optimizes an adversarial example with a set of transformed ({\it e.g.}, translated, resized) images, and has been proven effective to prevent the adversarial examples from overfitting to the white-box model being attacked \cite{xie2019improving,dong2019evading}, CAC fixes the top-{\it n} proposals of the region proposal network \cite{ren2015faster} and attacks all the classifications in the dense proposals, which significantly improve the transferability of the adversarial objects. Moreover, CAC can generate camouflages for arbitrary objects.

Further, to fairly evaluate the effectiveness of different attacks, we use the Unity simulation engine to build a photo-realistic 3D urban scene with high fidelity streets, buildings, plants, {\it etc}. The simulation engine enables us to conduct experiments under a variety of environmental conditions: lighting, backgrounds, camera-to-object distances, view angles, {\it etc}. Experimental results demonstrate that the 3D coated camouflage can consistently mislead the detectors from any viewpoint, which is superior to the piecing together patches. 

Figure \ref{fig:1} shows a sample of the generated adversarial vehicles in the physical world. We can fabricate the adversarial object by 3D printing, or simply print the adversarial texture with a color printer and paste it onto the original object. In summary, our main contributions are as follows:

\begin{itemize}
\item{To the best of our knowledge, our work is the first to learn coated adversarial camouflages, which are trained in 3D space and can cause object detectors to misidentify objects as designated target labels from arbitrary viewpoints.}

\item{We propose the dense proposals attack strategy to guarantee the attacking ability of the generated adversarial camouflages, especially improving the transferability of the adversarial objects under the black-box setting.}

\item{We build a Unity simulation scene to fairly and reproducibly evaluate the effectiveness of different attacks, and extensive experiments show that CAC achieves state-of-the-art results.}

\item{The proposed CAC can generate camouflages for any object and exhibits good generalization to the real world. In particular, an adversarial object can be fabricated by 3D printing directly, or obtained by pasting the camouflage to the surface of the original object.}
\end{itemize}

\section{Related Works}
\subsection{Digital Attacks}

Digital attacks generate adversarial noises for inputs in the digital pixel domain. A series of attack methods \cite{szegedy2014intriguing,xie2019improving,dong2019evading} have been proposed to generate adversarial examples to attack the image classifiers. Xie {\it et al.} \shortcite{xie2017adversarial} extended adversarial examples from image classification to object detection, and proposed the Dense Adversary Generation (DAG) method to generate visually imperceptible perturbations to fool detectors. DAG makes the proposals very dense by increasing the original threshold of non-maximal suppression (NMS) ({\it e.g.} from 0.7 to 0.9) in the first stage of Faster R-CNN \cite{ren2015faster}, thus the proposals on each image increase from 300 to around 3000. In contrast, CAC fixes the original top-$n$ ranked proposals after NMS, and attacks the dense classifications simultaneously in the second stage of Faster R-CNN. The noises of digital attacks are usually too subtle to be effective in the physical world due to the destructive environmental noises and input transformations \cite{lu2017no}.

\begin{figure*}[!ht]
\centering
\includegraphics[width=0.94\textwidth]{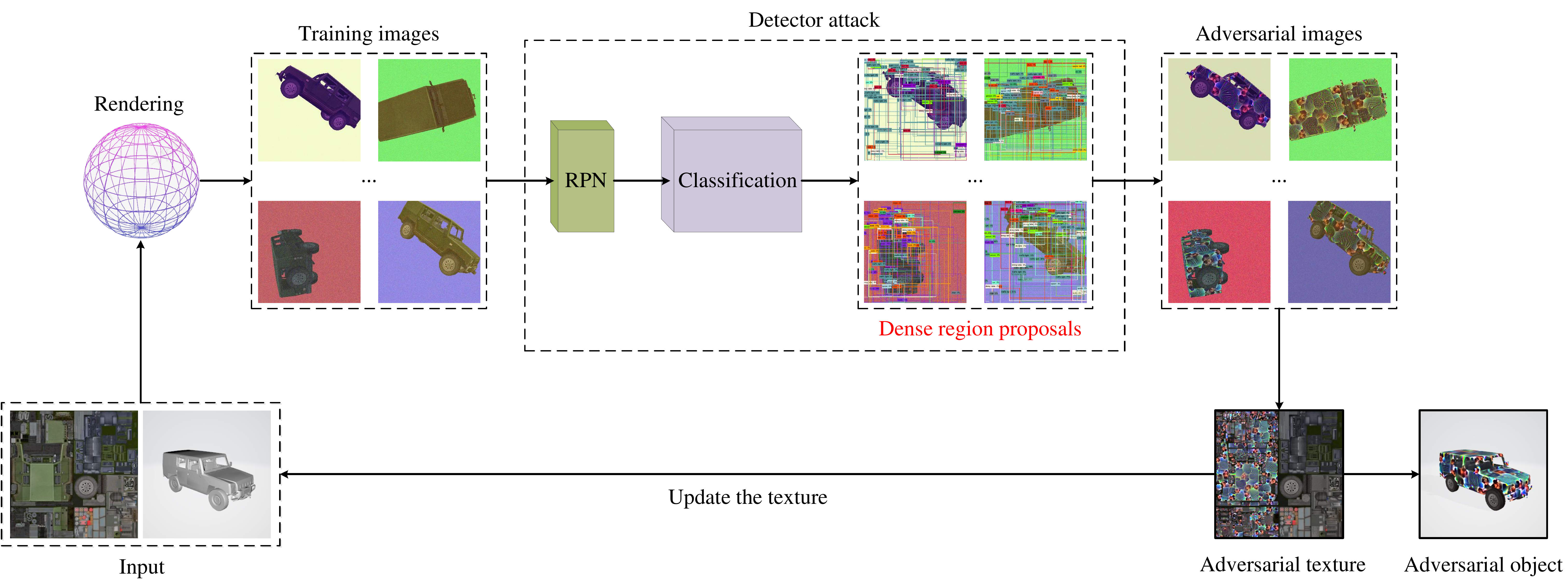} 
\caption{Overview of the pipeline to generate the adversarial camouflage texture. (1) Input transformation. The camouflage is mapped to a rendering of the vehicle, and the training images are obtained in real time by transforming viewpoints, lighting, backgrounds, printing errors, {\it etc.} according to a given distribution. (2) Detector attack. We simultaneously attack all the classifications in the fixed dense region proposals, and update the camouflage perturbations by minimizing the cross-entropy losses. Zoom in for more details.}
\label{fig:2}
\end{figure*}

\subsection{Physical Attacks}

Physical attacks usually add visible local noises to the input so that the generated adversarial examples remain adversarial in the physical world. Several works have studied attacks on detectors. Chen {\it et al.} \shortcite{chen2018shapeshifter} generated adversarial perturbed stop signs to fool detectors. Song {\it et al.} \shortcite{song2018physical} utilized synthetic transformations to attack object detection models, causing the object detectors to ignore the stop sign with sticker perturbations. Thys {\it et al.} \shortcite{thys2019fooling} and Huang {\it et al.} \shortcite{huang2020universal} learned adversarial patterns to attack instances belonging to the same object category. However, because these local adversarial patterns are trained in two-dimensional space, they would become less effective as the viewpoint changes in the 3D physical world. Zhang {\it et al.} \shortcite{zhang2019camou} generated mosaic-like full-coverage camouflages to make a vehicle randomly misidentified as other classes. In contrast, our goal is to make vehicles consistently misidentified as specific target classes closely related to traffic safety, such as person and stop sign, which would be more threatening and challenging.

\section{Methodology}

\subsection{Overview} 

We aim to generate camouflages that can fool the object detectors to misidentify the object as the target class or hide the object from being detected. We use the vehicle as an example to illustrate our method.

We simultaneously model both the object perspectives and the physical environment variations, so as to generate an adversarial object that is robust in the physical world. The transformation functions map the camouflage to a rendering of the vehicle, simulating functions including rotation, translation, perspective projection as well as lighting, background, printing errors and environmental noise changes.

Next, we attack the Faster R-CNN model \cite{ren2015faster}, a two-stage detector, under the white-box setting. We first run the region proposal network (RPN), and prune the region proposals by non-maximum suppression (NMS). Then we fix the top-$n$ pruned region proposals and feed them to the second stage classification, and attack all the classifications in these proposals simultaneously to generate robust camouflages \cite{ren2015faster,chen2018shapeshifter}. As illustrated in Figure \ref{fig:2}, CAC mainly consists of two steps:

\begin{itemize}
\item{Step 1. Obtaining the training images on-the-fly by simulating the 3D geometric transformations as well as the physical environment variations.}

\item{Step 2. Attacking all the classifications in the fixed dense region proposals simultaneously.}

\end{itemize}

\subsection{3D Rendering}
Let $(\textbf{m},\textbf{c})$ be a 3D object with a mesh tensor $\textbf{m}$ and a texture tensor $\textbf{c}$. The training image $x$ with ground-truth label $y$ is the rendered result of the 3D object $(\textbf{m},\textbf{c})$ with different environment conditions $e \in \textbf{E}$  ({\it e.g.}, view angles, distances, backgrounds, printing errors, {\it etc}., the distribution detailed in Table 3 in the Appendix) from a renderer $\mathcal{R}$ by
\begin{equation}
x=\mathcal{R}((\textbf{m},\textbf{c}),e)
\end{equation}

We obtain the adversarial camouflage $\textbf{c}_{adv}$ by adding perturbations to $\textbf{c}$ and generate the adversarial example as 
\begin{equation}
x_{adv}=\mathcal{R}((\textbf{m},\textbf{c}_{adv}),e)
\end{equation}
where $(\textbf{m},\textbf{c}_{adv})$ is the obtained 3D adversarial object. Different from the 2D local adversarial patches, $\textbf{c}_{adv}$ performs corresponding spatial transformations according to the pose changes of the object during the training process.

\subsection{Dense Proposals Attack}
The Faster R-CNN is a two-stage model. The region proposal network (RPN) in the first stage is a fully convolutional network that simultaneously predicts object bounds and scores at each position. The classifier in the second stage performs classification in each region proposal. Each detection includes a probability distribution over $K$ pre-defined classes as well as the location of the detected object. 

For the targeted attack, it aims to fool the deep model $f( \cdot )$ into outputting a specific target label $y^*$, which can be express as $f(x_{adv}) = y^*$, and $y^* \ne y$. The objective is to minimize the cross-entropy loss function $J(f(x_{adv}),y^*)$ of the classifier. We do not constrain the distance between the ${\textbf{c}}_{adv}$ and the original $\textbf{c}$, because for three-dimensional objects, the textures hardly impede human recognition. Therefore, the optimization problem for attacking the classification of an object in one proposal can be written as 
\begin{equation}
\mathop {\arg \min }\limits_{{{\textbf{c}}_{adv}}} J(f(x_{adv}),y^*)
\label{Eq3}
\end{equation}

Some RPN proposals highly overlap with each other. To reduce redundancy, most models adopt non-maximum suppression (NMS) to prune the proposal regions based on their confidence scores \cite{ren2015faster}. To improve the attacks, in each iteration, we first run the region proposal network, then fix the top-$n$ pruned proposals.  The label of each proposal is defined as the corresponding confident class. In the second classification stage, we minimize the cross-entropy losses between the target class and the predicted classes in all the fixed dense proposals. Similar to data augmentation, the objects in the fixed region proposals can be regarded as a set of transformed ({\it e.g.}, translated, cropped) sub-images, which can alleviate the overfitting phenomena and improve the transferability of the generated adversarial objects.

\begin{algorithm}[!t]
\caption{Algorithm of CAC}
\label{alg:1}
\textbf{Input}: 3D object $(\textbf{m},\textbf{c})$, environment condition parameter $e \in \textbf{E}$, neural renderer $\mathcal{R}$, target class label $y^*$, detector $f$, maximal iteration number $N$.\\
\textbf{Output}: Adversarial camouflage tensor $\textbf{c}_{adv}$.

\begin{algorithmic}[1] 
\STATE $\textbf{c}_{adv}^0 \leftarrow  \textbf{c}$;
\FOR {$t = 0$ to $ N-1 $}
\STATE Generate training images in each iteration:\\ $x_{adv}^t \leftarrow \mathcal{R}((\textbf{m},\textbf{c}_{adv}^t),e)$ 
\STATE Obtain the dense region proposals: \\$\mathcal{P}=\{p_i|p_i=(s_i, b_i); i=1,2,3 ... n\}$;
\STATE Update $\textbf{c}_{adv}^t$ via attacking all the classifications of the proposals :\\
$\mathop {\arg \min }\limits_{{{\textbf{c}}_{adv}^t}} \mathbb{E}{_{{x \sim {\bf{X}}},{e \sim {\bf{E}}}}}[\frac{1}{n}\sum\limits_{{p_i} \in \mathcal{P}} {J(f(x_{adv}^t,p_i),{y^*})} ] $,\\
${{\textbf{c}}_{adv}^t}= {\rm {Clip}} ({{\textbf{c}}_{adv}^t},0,1)$;
\ENDFOR
\STATE \textbf{return}: $\textbf{c}_{adv} = \textbf{c}_{adv}^N$.
\end{algorithmic}
\end{algorithm}

Let $n$ be the number of the fixed proposals, and the output proposals of each image is $\mathcal{P}=\{p_i|p_i=(s_i, b_i); i=1,2,3 ... n\}$, where $s_i$ is the confidence score and $b_i$ represents the location of the $i$-th region proposal \cite{huang2020universal}. In order to enhance the attack, rather than optimize the objective function at a single point as Eq. (\ref{Eq3}), CAC simultaneously attacks the classifications of all the fixed dense region proposals to optimize the adversarial camouflage as 

\begin{equation}
\mathop {\arg \min }\limits_{{{\textbf{c}}_{adv}}} [\frac{1}{n}\sum\limits_{{p_i} \in \mathcal{P}} {J(f(x_{adv},p_i),{y^*})} ] 
\end{equation}

Therefore, the camouflage is trained to optimize the object function 
\begin{equation}
\mathop {\arg \min }\limits_{{{\textbf{c}}_{adv}}} \mathbb{E}{_{{x \sim {\bf{X}}},{e \sim {\bf{E}}}}}[\frac{1}{n}\sum\limits_{{p_i} \in \mathcal{P}} {J(f(x_{adv},p_i),{y^*})} ] 
\end{equation}
where $\bf{X}$ is the training set of images generated in real time by the renderer, $\bf{E}$ is the distribution of the environment conditions simulated by the renderer, and $\mathbb{E}$ is the Expectation over Transformation (EOT) technique \cite{athalye2018synthesizing} which models the adversarial perturbations within the optimization procedure. The ``true" input $\mathcal{R}((\textbf{m},\textbf{c}_{adv}),e)$ perceived by the detector $f( \cdot )$ is the input object $(\textbf{m},\textbf{c}_{adv})$ with environment condition $e$ after the render process. It optimizes the losses between the expected detection results and the target class $y^*$. The resultant camouflage pixel value is clipped to the valid range ({\it i.e.}, [0,1] for images). The overall procedure of CAC is summarized in Algorithm \ref{alg:1}.

\section{Experiments}

\subsection{Experimental Settings}
\paragraph{Source Model.} We generate adversarial camouflage on the Faster R-CNN with Inception-v2 \cite{szegedy2016rethinking} as the backbone network. The model is trained on the COCO2014 dataset \cite{lin2014microsoft}. We denote this model as FR-Incv2-14. Faster R-CNN adopts a two-stage detection strategy, the first stage generates many region proposals that may contain objects, and the second stage outputs the classification results and the refined bounding box coordinates.

\paragraph{Target Models.} To evaluate the cross-model and cross-training transferability of the adversarial camouflage generated by the source model, we test the results on two-stage and one-stage detectors with different network structures and training datasets. For two-stage detectors, we evaluate the performance of the Faster R-CNN model, in addition to Inception-v2, the backbone networks include VGG16 \cite{simonyan2014very} and ResNet101 \cite{he2016deep}, which are either trained on the PascalVOC-2007 trainval set or the combined PascalVOC-2007 and PascalVOC-2012 trainval set \cite{everingham2015pascal}, or on COCO2014. We denote these models as FR-VGG16-07, FR-RES101-07, FR-VGG16-0712, FR-RES101-0712, FR-VGG16-14 and FR-RES101-14, respectively. For one-stage detectors, we select YOLOv3 \cite{redmon2018yolov3} and YOLOv5 \footnote{\url{https://github.com/ultralytics/yolov5}}, whose network structures are quite different from Faster R-CNN, making it more challenging to conduct transfer attacks, and we denote these models trained on COCO2014 as YOLOv3-14 and YOLOv5-14, respectively.

\paragraph{Evaluation Metrics.} The principal quantitative measure of the detection task is the average precision. Detections are considered true or false positives based on the area Intersection over Union (IoU) between the predicted box $B_p$ and the ground truth box $B_{gt}$, which is defined as $IoU=\frac{{{B_p} \cap {B_{gt}}}}{{{B_p} \cup {B_{gt}}}}$. The threshold of IoU is set to 0.5 as in the PASCAL VOC detection challenge  \cite{everingham2015pascal} to determine whether the detector hits or misses the true category. This metrics is denoted as P@0.5 \cite{zhang2019camou,huang2020universal}. The confidence score threshold of all models is set to 0.3 for evaluation. (Since the jeep vehicle with the original texture is detected as a truck or a car, we consider both classes to be ``true" classes).

\paragraph{Baselines.} We generate three simple textures for comparison, {\it Natural}, {\it Naive} and {\it Random}. In addition, we compare our method to several state-of-the-art physical attacks. Adversarial patch ({\it AdvPat}) \cite{thys2019fooling} generates adversarial patches to fool a detector by minimizing different probabilities related to the appearance of a object. Shapeshifter ({\it Shape}) \cite{chen2018shapeshifter} obtains adversarial objects that mislead a detector via EOT technique \cite{athalye2018synthesizing}. {\it UPC} \cite{huang2020universal} crafts adversarial camouflages for non-rigid or non-planar objects. {\it Camou} \cite{zhang2019camou} learns camouflages that can hide a vehicle from detectors. For fair comparison, the adversarial patterns generated by {\it AdvPat}, {\it Shape} and {\it UPC} are pasted on all sides of the vehicles. We restrict the CAC camouflage to the body of the vehicle, leaving wheels, windows, lightings, {\it etc}. unaltered as the discriminative visual cues for the detectors. Nine different textures are shown in Figure 7 in the Appendix.

\paragraph{Physical Simulation.} To fairly and reproducibly evaluate different attacks, we build a photo-realistic 3D simulation scene (see Figure 8 in the Appendix).

\begin{figure}[!t]
\centering
\includegraphics[width=0.95\columnwidth]{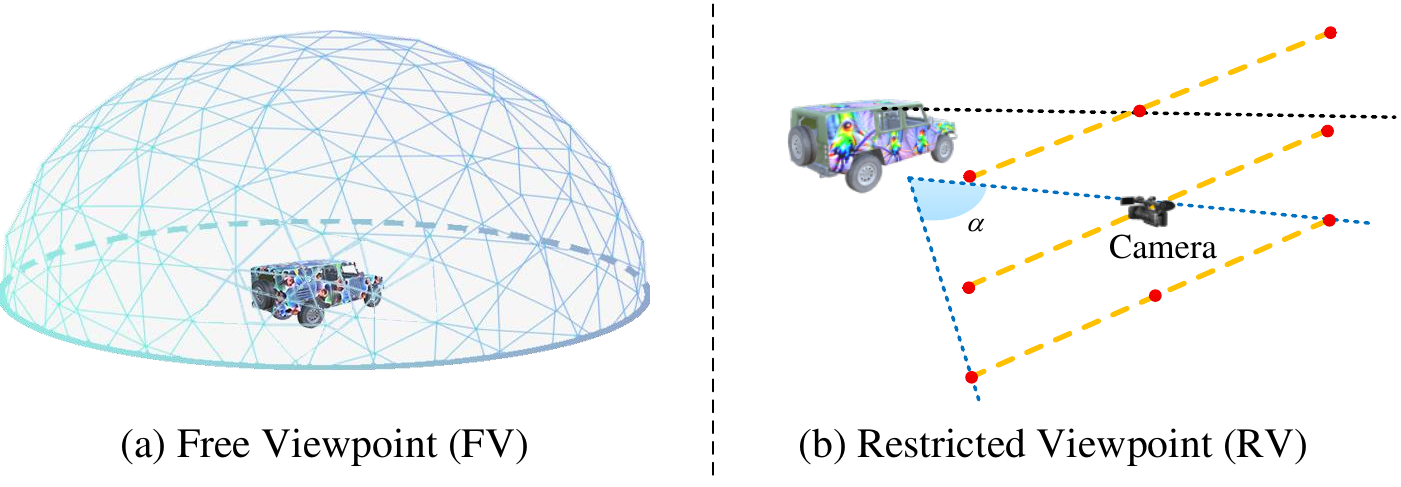} 
\caption{The camera settings. (a) The camera is placed arbitrarily within a semi-ellipsoid as free viewpoint; (b) the camera is restricted to a certain range of view angles ({\it e.g.}, {\it UPC} with restricted view angle range $\alpha$ in its original paper).}
\label{fig:4}
\end{figure}

\subsection{Virtual Scene Experiments}
The area light and directional light sources are used to get 2 levels of brightness ({\it i.e.}, bright:$L_b$ and dark:$L_d$) simulation scenes (illustrated in Figure \ref{fig:1} (a)).

As illustrated in Figure \ref{fig:4} (a), the view angles and distances of the cameras distribute freely as {\it Free Viewpoint (FV)} within a semi-ellipsoid, so that the attack effectiveness can be evaluated more accurately and comprehensively.
 
We rendered 400 ($200 \times 2$) images with free viewpoints in different locations of the simulation scene for each of the nine vehicles with different textures. The target class of CAC attack is ``person".

\subsubsection{Denseness of Proposals}
\begin{figure}[!t]
\centering
\includegraphics[width=0.85\columnwidth]{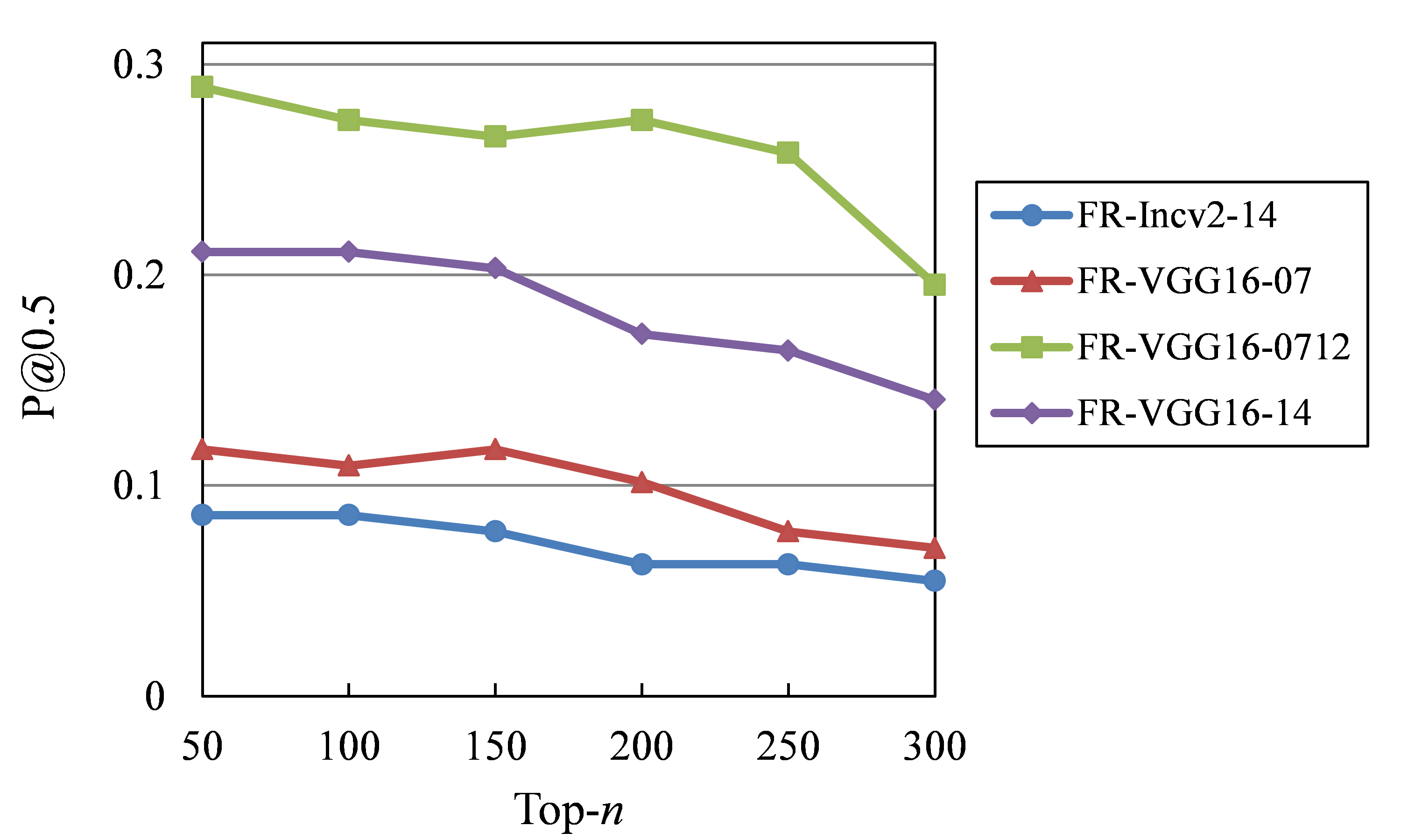} 
\caption{The average precision P@0.5 of using adversarial camouflages generated on FR-Incv2-14 to attack FR-VGG16-07, FR-VGG16-0712 and FR-VGG16-14, with respect to the top-$n$ number.}
\label{fig:5}
\end{figure}

We first study the impact of the proposal denseness on attacks. Different number of proposals are fixed after the RPN, ranging from 50 to 300. As can be seen in Figure \ref{fig:5}, the average precision goes down as the top-$n$ number increases, which indicates that denser proposals obtain stronger adversarial camouflages. Therefore, we choose a large number (300) that performs better.

\subsubsection{Comparison with Simple Textures} 
The average precision results for models trained on the COCO2014 dataset are shown in Table \ref{tab:1}, and for models trained on PascalVOC datasets are shown in Table 4 in the Appendix due to the space limitations. We can find that for the Natural/Naive/Random textures, the average precisions of almost all models drop little, indicating that simple textures hardly affect the detectors, {\it i.e.}, the detectors are robust to the simulation data, even for the noisy random texture. In contrast, the camouflage generated by CAC significantly reduces the average precisions under both the bright and dark brightness conditions, which verifies the effectiveness of the attack. 

In addition, we can observe that for the simple textures and the adversarial textures, most of the attacks in the dark environment are stronger than those in the bright environment, which may be attributed to the fact that the robustness of the detector is relatively weak when the brightness level is low.

\subsubsection{Comparison with Existing Attacks} 
Except for the {\it Camou}, the adversarial patterns of {\it AdvPat}, {\it Shape} and {\it UPC} are local patches on different sides of the vehicles (see Figure 7 in the Appendix). The patches are trained in 2D space and evaluated from a restricted view ({\it e.g.}, side view) in their original papers. We denote the side view as {\it Restricted Viewpoint (RV)} (shown in Figure \ref{fig:4} (b), $\alpha$ is set to ${120^ \circ }$, {\it i.e.}, the left and right view angles are within ${60^ \circ }$). However, the cameras in the wild do not film the vehicle object only from the side, so we evaluate the attacks under both the free viewpoint and restricted viewpoint settings.

Table \ref{tab:3} shows the physical attack results for models trained on the COCO2014 dataset, and the results for models trained on PascalVOC datasets are shown in Table 5 in the Appendix. It can be found that the local adversarial patterns can also reduce the average precision when pasted on other sides of the vehicles. However, the adversarial camouflage generated by CAC exhibits significantly better performance, and has much higher drop rates both in the restricted and free viewpoints than the baseline methods. The results demonstrate that the attack effectiveness of our coated camouflage outperforms the piecing-together patches. Besides, CAC is also superior to the full-coverage {\it Camou} texture, which further demonstrates the cross-model transferability of CAC. It worth to note that {\it Camou} conducts non-targeted attacks, while CAC conducts targeted attacks, which are more challenging.

We find that the specific target class label is well transferred, {\it i.e.}, the target models also misidentify the vehicles with our adversarial camouflage trained on the source model as the person class label. This is because different models have similar decision boundaries due to the similar or same training dataset \cite{xie2019improving}. See Figures 9 to 12 in the Appendix for more details.

Moreover, it can be seen that the average precision in restricted viewpoint is usually higher than that in free viewpoint ({\it i.e.}, lower drop rates), indicating that the attacks are more difficult to succeed from the side view.

\begin{table*}[!htb]
\centering
\resizebox{\textwidth}{12mm}{
\begin{tabular}{c|ccc|ccc|ccc|ccc|ccc}
\hline
Model      & \multicolumn{3}{c|}{FR-Incv2-14} & \multicolumn{3}{c|}{FR-VGG16-14} & \multicolumn{3}{c|}{FR-RES101-14} & \multicolumn{3}{c|}{YOLOv3-14} & \multicolumn{3}{c}{YOLOv5-14}  \\ \hline
Scheme     & $L_b$      & $L_d$      & Avg (Drop)    & $L_b$      & $L_d$      & Avg (Drop)    & $L_b$      & $L_d$      & Avg (Drop)     & $L_b$       & $L_d$      & Avg (Drop)     & $L_b$       & $L_d$       & Avg (Drop)        \\ \hline
Original   & 1.00    & 1.00    & 1.00 (-)      & 0.96    & 0.98    & 0.97 (-)       & 0.98     & 0.99    & 0.98 (-)      & 1.00    & 1.00    & 1.00 (-)       & 1.00    & 1.00     & 1.00 (-)           \\
Natural    & 1.00    & 1.00    & 1.00 (0)      & 0.90    & 0.92    & 0.91 (0.06)    & 0.91     & 0.93    & 0.92 (0.06)   & 0.93    & 0.90    & 0.92 (0.08)    & 0.93     & 0.96     & 0.95 (0.05)      \\
Naive      & 1.00    & 0.94    & 0.97 (0.03)   & 0.93    & 0.83    & 0.88 (0.09)    & 0.92     & 0.94    & 0.93 (0.05)   & 0.93    & 0.86    & 0.90 (0.10)    & 0.92     & 0.95     & 0.94 (0.06)       \\
Random     & 0.90    & 0.93    & 0.91 (0.09)   & 0.86   & 0.84    & 0.85 (0.12)    & 0.92     & 0.90    & 0.91 (0.07)   & 0.84    & 0.76    & 0.80 (0.20)    & 0.90     & 0.88     & 0.89 (0.11)      \\
CAC & 0.08$^*$    & 0.04$^*$    & 0.06 (\textbf{0.94})   & 0.18    & 0.10    & 0.14 (\textbf{0.83})    & 0.59     & 0.47    & 0.53 (\textbf{0.45})   & 0.21    & 0.25 & 0.23 (\textbf{0.77})    & 0.54 & 0.44     & 0.49 (\textbf{0.51})      \\ \hline
\end{tabular}}
\caption{Average precision P@0.5 and drop rates in virtual scene experiments under two illumination level settings. Each P@0.5 is averaged over free viewpoints. * indicates the white-box attacks. The best results are in bold.}
\label{tab:1}
\end{table*}

\begin{table*}[!htb]
\centering
\resizebox{\textwidth}{14mm}{
\begin{tabular}{c|cc|cc|cc|cc|cc}
\hline
Model     & \multicolumn{2}{c|}{FR-Incv2-14} & \multicolumn{2}{c|}{FR-VGG16-14} & \multicolumn{2}{c|}{FR-RES101-14} & \multicolumn{2}{c|}{YOLOv3-14} & \multicolumn{2}{c}{YOLOv5-14}  \\ \hline
Scheme    & RV (Drop)       & FV (Drop)      & RV (Drop)      & FV (Drop)      & RV (Drop)       & FV (Drop)       & RV (Drop)        & FV (Drop)       & RV (Drop)        & FV (Drop)              \\ \hline
Original  & 1.00 (-)        & 1.00 (-)       & 1.00 (-)       & 0.97 (-)        & 1.00 (-)         & 0.98 (-)       & 1.00 (-)        & 1.00 (-)        & 1.00 (-)         & 1.00 (-)           \\
AdvPat  & 0.92 (0.08)     & 0.82 (0.18)    & 0.97 (0.03)    & 0.73 (0.24)     & 0.99 (0.01)      & 0.77 (0.21)    & 0.33$^*$ (0.67)     & 0.30$^*$ (0.70)     & 0.70 (0.30)      & 0.59 (0.41)        \\
Shape   & 0.99$^*$ (0.01)       & 0.85$^*$ (0.15)   & 0.99 (0.01)    & 0.86 (0.11)     & 0.99 (0.01)      & 0.85 (0.13)    & 0.95 (0.05)     & 0.74 (0.26)     & 0.96 (0.04)         & 0.78 (0.22)         \\
UPC     & 0.95 (0.05)     & 0.74 (0.26)    & 0.79 (0.21)    & 0.56 (0.41)     & 0.93 (0.07)     & 0.82 (0.16)    & 0.32 (0.68)     & 0.34 (0.66)    & 0.69 (0.31)      & 0.57 (0.43)     \\
Camou   & 0.97 (0.03)     & 0.55 (0.45)    & 0.94 (0.06)    & 0.52 (0.45)     & 0.99 (0.01)      & 0.69 (0.29)    & 0.51$^*$ (0.49)        & 0.28$^*$ (0.72)     & 0.94 (0.06)      & 0.64 (0.36)      \\
CAC & 0.08$^*$ (\textbf{0.92})    & 0.06$^*$ (\textbf{0.94})   & 0.22 (\textbf{0.78})    & 0.14 (\textbf{0.83})     & 0.85 (\textbf{0.15})      & 0.53 (\textbf{0.45})    & 0.29 (\textbf{0.71})     & 0.23 (\textbf{0.77})     & 0.67 (\textbf{0.33})      & 0.49 (\textbf{0.51})       \\ \hline
\end{tabular}}
\caption{Average precision P@0.5 and drop rates of different attacks in restricted and free viewpoints. Each P@0.5 is averaged over two brightness levels. * indicates the white-box attacks. The best results are in bold.}
\label{tab:3}
\end{table*}

\subsection{Physical Environment Experiments}
We evaluate physical attacks in the real world over a 3D-printed vehicle with the adversarial camouflage generated by CAC. We take photos and record videos with an HONOR 20 cell phone. Similar to the virtual scene experiments, the 3D-printed vehicle is filmed in free viewpoints. We put the vehicle on a rotating turntable to eliminate blind spots. Figure 14 shows some qualitative results, and a demo video can be found at: https://www.bilibili.com/video/BV1zL411J73r.

\begin{figure}[!t]
\centering
\includegraphics[width=0.85\columnwidth]{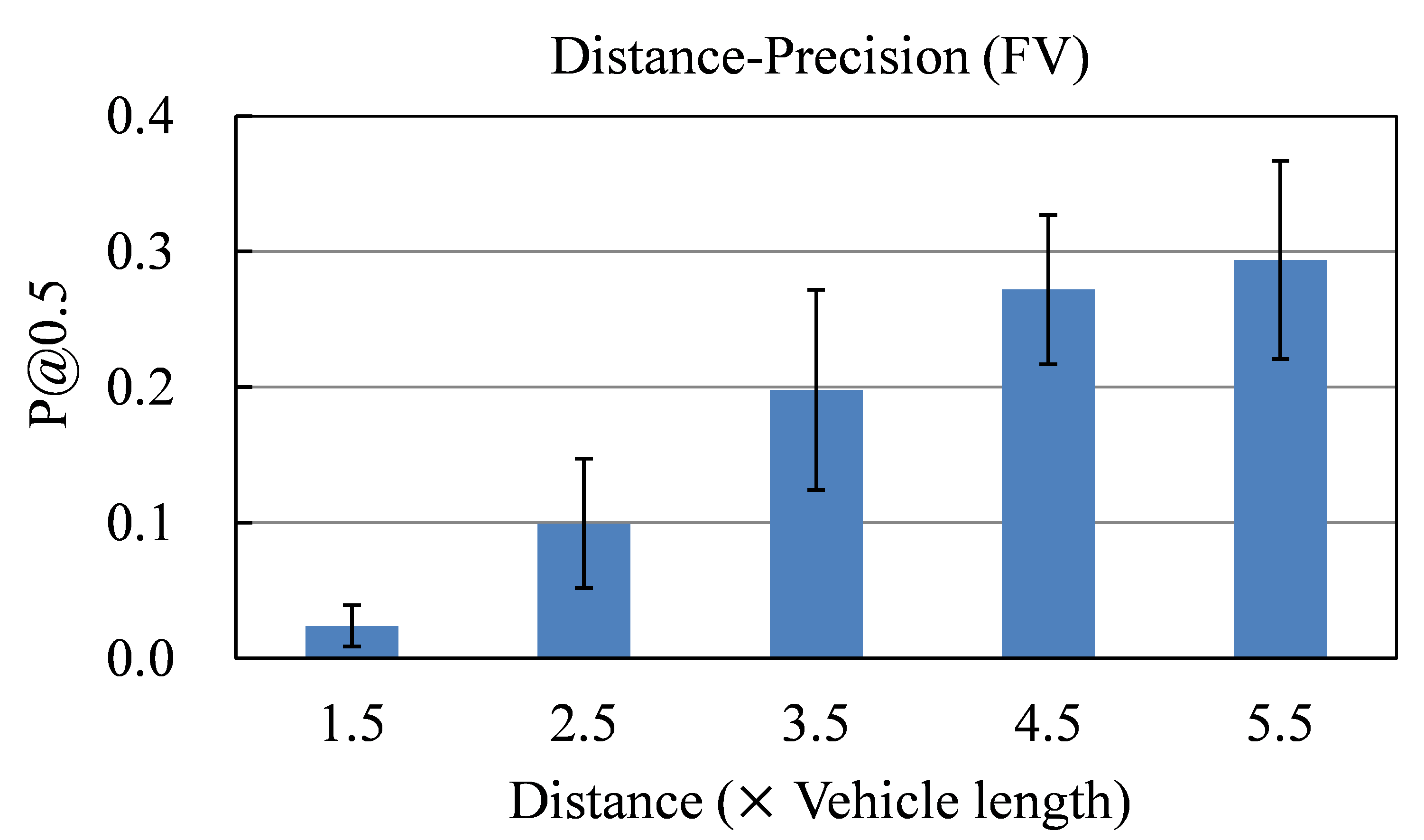} 
\caption{The P@0.5 (with $ \pm std$ over 5 tests) under different distance conditions in the real world.}
\label{fig:6}
\end{figure}

In Figure \ref{fig:6} we plot the relationship between the camera-to-object distance and average precision of FR-Incv2-14 under the $360^\circ$ free viewpoint setting. The abscissa is the ratio of distance to the vehicle length. It can be seen that the average precision goes up as the distance increases, this is probably because the camouflage is captured with lower quality from a distance. The results show low detection precisions, which demonstrate that the 3D-printed adversarial vehicle is strongly adversarial over a variety of viewpoints, and the camouflage generated by the virtual scene experiments exhibits high generalization to the real world.

\subsection{Model Attention Analysis}
\begin{figure}[!t]
\centering
\includegraphics[width=0.9\columnwidth]{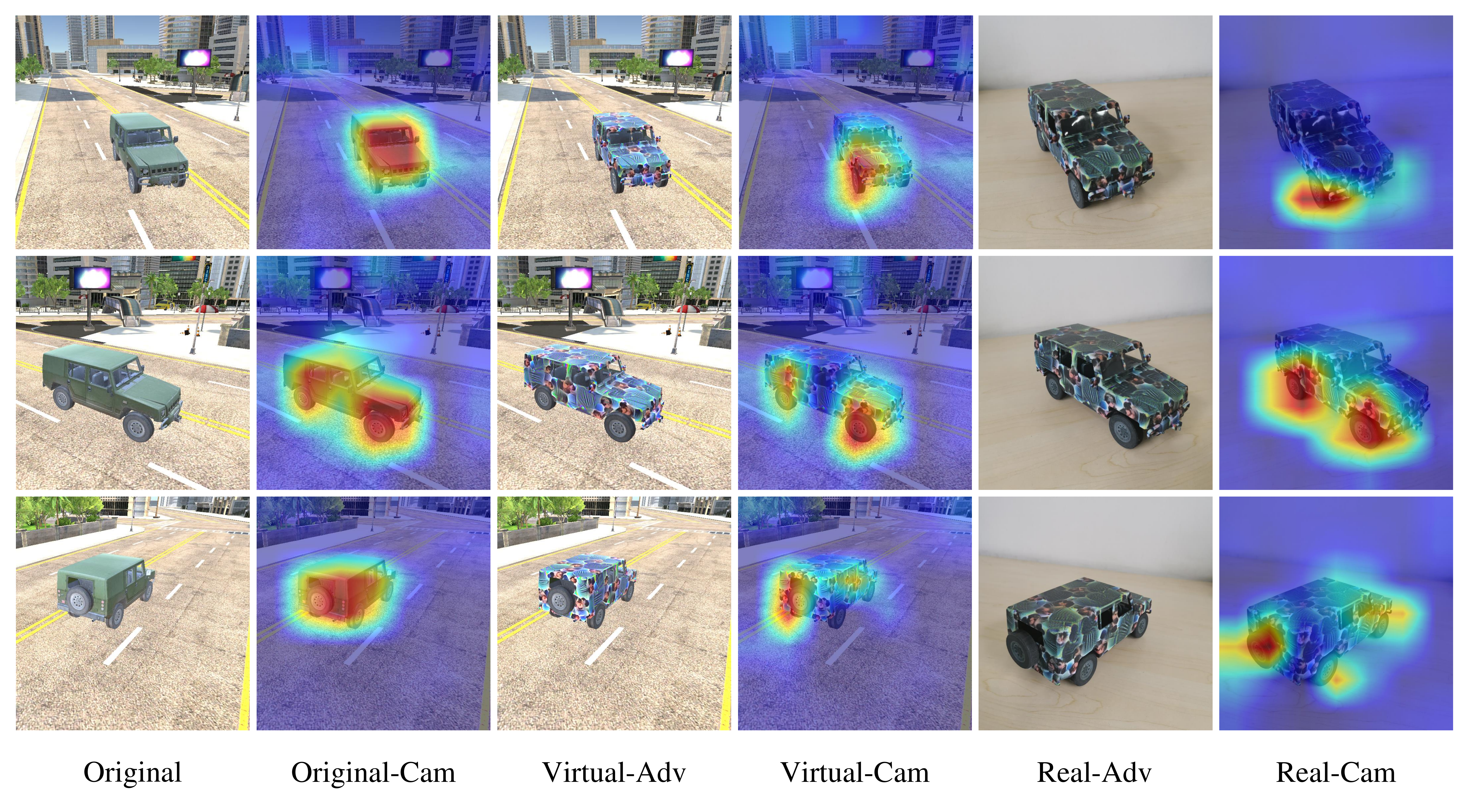} 
\caption{Visualization of discriminative regions for ``jeep" class label in the virtual scene and the real world. For the vehicles with the adversarial camouflage, the model places its attention predominantly on the regions with no adversarial perturbations ({\it e.g.}, wheels).}
\label{fig:7}
\end{figure}

The regions that the models pay attention to can be deemed as the discriminative regions. We generate the attention maps of the vehicle with different viewpoints on VGG16 model by the model-agnostic Grad-CAM \cite{selvaraju2017grad} technique. Figure \ref{fig:7} shows the original vehicle, virtual adversarial vehicle (Virtual-Adv), 3D printed adversarial vehicle (Real-Adv) and their attention maps for the ``jeep" class label, respectively. We can observe the CAC attack distracts the attention maps from the vehicle body and focuses them in the wheel areas where there is no adversarial noise. This also explains why restricted side-view attacks are more difficult than free-view attacks in Table \ref{tab:3}, as the noiseless wheels are a very salient vehicle distinguishing feature in the side view.

\subsection{Attacks for Other Objects and Classes} 
To demonstrate the generalization of CAC, we generate adversarial camouflages to fool other objects. As shown in Figure 15 in the Appendix, the barrels and containers with adversarial camouflages can be misidentified as the target label in different viewpoints, indicating that CAC is generic for any object beyond vehicles. In addition, we generate other target class ({\it e.g.}, stop sign) camouflages closely related to traffic safety for the vehicle, the qualitative results in Figure 16 in the Appendix further prove that CAC can fool the object detector into outputting arbitrary specific class labels.

\section{Conclusion}
In this paper, we investigate the problem of generating robust 3D adversarial camouflages in the physical world for object detectors. By modeling the 3D rendering, and using a set of dense proposals to optimize the adversarial camouflage in each iteration, the vehicle with the resultant adversarial camouflage generated by the proposed CAC can fool object detectors from any viewpoint, and exhibits significantly better performance than the state-of-the-art baseline methods. In addition, we build a 3D scene to fairly and reproducibly evaluate different attacks. With the 3D printing technology, we successfully fabricate the first physical adversarial object that is detected as a specific target class under arbitrary viewpoints and different lighting conditions. For future work, we plan to make the camouflages visually more natural.

\bibliographystyle{named}
\bibliography{ijcai22}

\appendix

\section*{Appendix}

\section{Environment Condition}

As shown in Table \ref{tab:3?}, we approximately model the physical-world changes and printing errors by a distribution of environment condition parameters, including 3D geometric transformations as well as the camera noise, additive and multiplicative lighting and per-channel color inaccuracies.

\renewcommand\thetable{3}   
\begin{table}[!htb]
\footnotesize
\centering
\begin{tabular}{lrr}
\hline
Environment condition           & Minimum       & Maximum       \\ \hline
Camera distance                 & 1.0           & 6.0           \\
X/Y translation                 & -0.05         & 0.05          \\
Rotation                        & \multicolumn{2}{c}{any}       \\
Background                      & (0.1,0.1,0.1) & (1.0,1.0,1.0) \\
Lighten / Darken (additive)       & -0.15         & 0.15          \\
Lighten / Darken (multiplicative) & 0.5           & 2.0           \\
Per-channel (additive)          & -0.15         & 0.15          \\
Per-channel (multiplicative)    & 0.7           & 1.3           \\
Gaussian Noise (stdev)          & 0.0           & 0.1           \\ \hline
\end{tabular}
\caption{Distribution of environment condition parameters for the physical world, approximating rendering, physical-world phenomena and
printing errors.}
\label{tab:3?}
\end{table}

\renewcommand\thefigure{7}  
\begin{figure}[!ht]
\centering
\includegraphics[width=0.9\columnwidth]{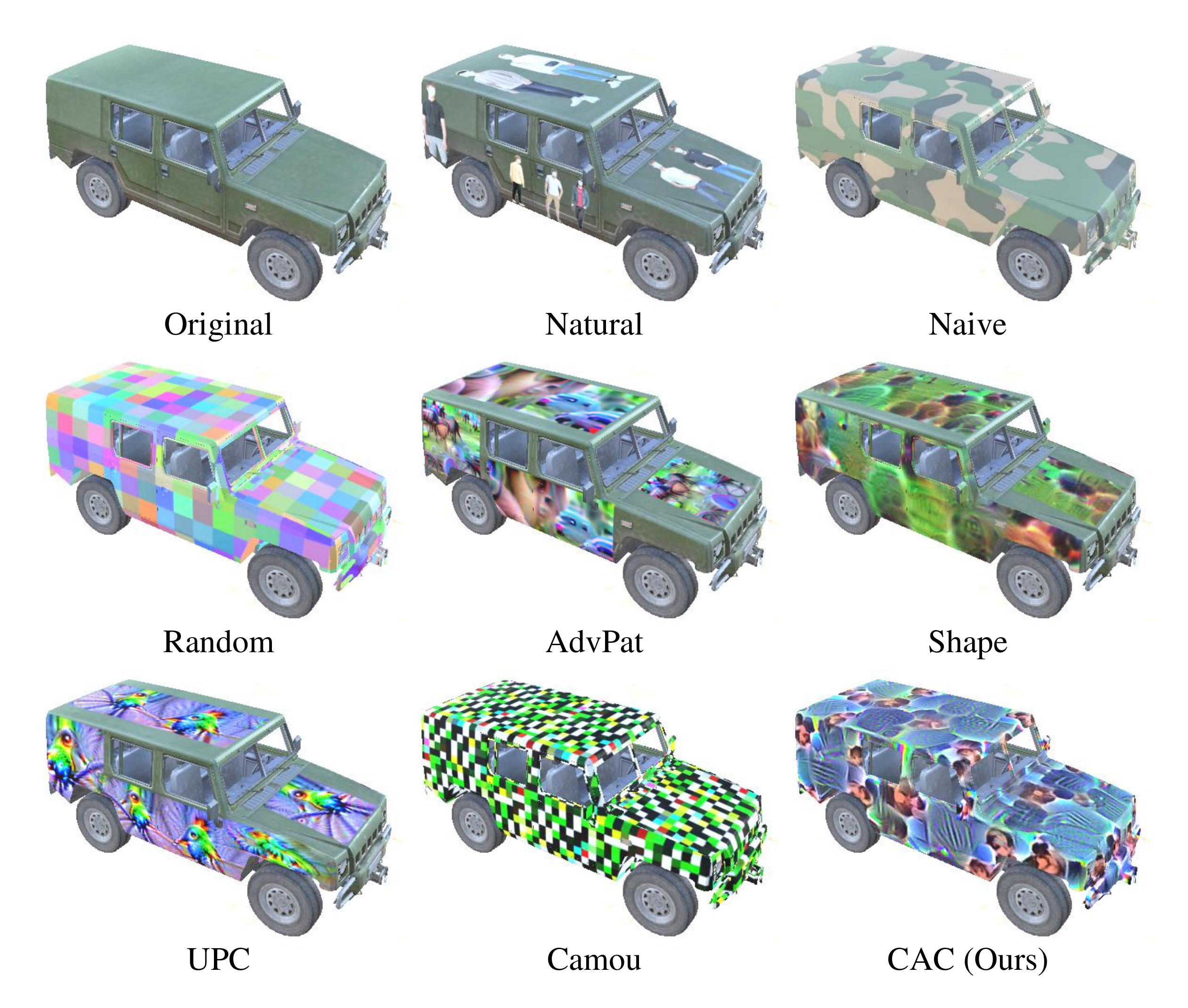} 
\caption{Examples of different textures. \textbf{Original}: vehicle with the original texture; \textbf{Natural}: vehicle with natural images as camouflage patterns; \textbf{Naive}: vehicle with simple army camouflage; \textbf{Random}: vehicle with random mosaic camouflage; \textbf{Advpat}, \textbf{Shape}, \textbf{UPC} and \textbf{Camou} are vehicles with adversarial patterns generated by the baseline methods; \textbf{CAC}: vehicle with the adversarial camouflage generated by the proposed CAC.}
\label{fig:3}
\end{figure}

\section{Virtual Scene}
Figure \ref{fig:3} shows vehicles with different textures. Figure \ref{fig:8} shows some locations of the virtual scene.

\renewcommand\thefigure{8}  
\begin{figure}[!ht]
\centering
\includegraphics[width=0.95\columnwidth]{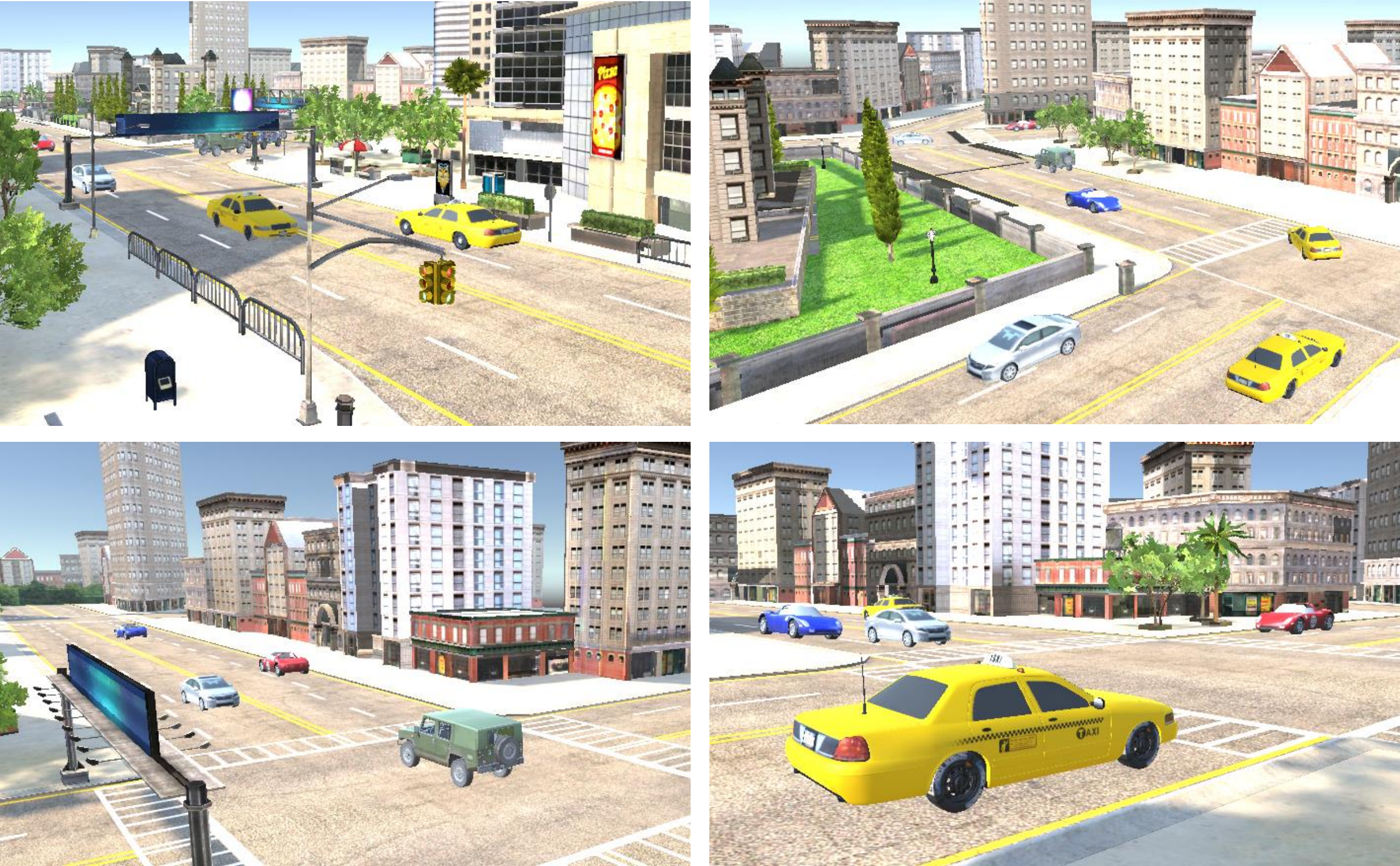} 
\caption{An urban scene built for evaluating the attacks.}
\label{fig:8}
\end{figure}

\renewcommand\thetable{4}   
\begin{table*}[!htb]
\centering
\footnotesize
\begin{tabular}{c|ccc|ccc|ccc|ccc}
\hline
Model      & \multicolumn{3}{c|}{FR-VGG16-07} & \multicolumn{3}{c|}{FR-VGG16-0712} & \multicolumn{3}{c|}{FR-RES101-07} & \multicolumn{3}{c}{FR-RES101-0712}   \\ \hline
Scheme     & $L_b$      & $L_d$      & Avg (Drop)    & $L_b$      & $L_d$      & Avg (Drop)    & $L_b$      & $L_d$      & Avg (Drop)     & $L_b$       & $L_d$      & Avg (Drop)       \\ \hline
Original   & 0.93    & 0.97    & 0.95 (-)       & 0.96     & 0.97    & 0.97 (-)      & 0.98    & 0.99    & 0.99 (-)       & 0.99    & 0.99     & 0.99 (-)           \\
Natural    & 0.89    & 0.91    & 0.90 (0.05)      & 0.90    & 0.94    & 0.92 (0.05)    & 0.95     & 0.99    & 0.97 (0.02)   & 0.94    & 0.96    & 0.95 (0.04)    \\
Naive      & 0.93    & 0.84    & 0.89 (0.06)   & 0.94    & 0.86    & 0.90 (0.07)    & 0.92     & 0.91    & 0.92 (0.07)   & 0.90    & 0.92    & 0.91 (0.08)     \\
Random     & 0.84    & 0.80    & 0.82 (0.13)   & 0.90   & 0.87    & 0.89 (0.08)    & 0.89     & 0.90    & 0.90 (0.09)   & 0.96    & 0.94    & 0.95 (0.04)    \\
CAC & 0.07    & 0.06   & 0.07 (\textbf{0.88})   & 0.21    & 0.17    & 0.19 (\textbf{0.78})    & 0.51    & 0.45    & 0.48 (\textbf{0.51})   & 0.68    & 0.58 & 0.63 (\textbf{0.36})         \\ \hline
\end{tabular}
\caption{Average precision P@0.5 and drop rates in virtual scene experiments under two illumination level settings. Each P@0.5 is averaged over free viewpoints. The best results are in bold.}
\label{tab:2}
\end{table*}

\renewcommand\thetable{5}   
\begin{table*}[!htb]
\centering
\footnotesize
\begin{tabular}{c|cc|cc|cc|cc}
\hline
Model     & \multicolumn{2}{c|}{FR-VGG16-07} & \multicolumn{2}{c|}{FR-VGG16-0712} & \multicolumn{2}{c|}{FR-RES101-07} & \multicolumn{2}{c}{FR-RES101-0712}  \\ \hline
Scheme    & RV (Drop)       & FV (Drop)      & RV (Drop)      & FV (Drop)      & RV (Drop)       & FV (Drop)       & RV (Drop)        & FV (Drop)        \\ \hline
Original  & 1.00 (-)        & 0.95 (-)       & 1.00 (-)       & 0.97 (-)        & 1.00 (-)         & 0.99 (-)       & 1.00 (-)        & 0.99 (-)                \\
AdvPat  & 0.40 (0.60)     & 0.29 (0.66)    & 0.47 (0.53)    & 0.44 (0.53)     & 0.89 (0.11)      & 0.66 (0.33)    & 0.94 (0.06)     & 0.81 (0.18)      \\
Shape   & 0.74 (0.26)       & 0.62 (0.33)   & 0.81 (0.19)    & 0.73 (0.24)     & 0.97 (0.03)      & 0.83 (0.16)    & 0.99 (0.01)     & 0.93 (0.06)              \\
UPC     & 0.11 (0.89)     & 0.10 (0.85)    & 0.09$^*$ (\textbf{0.91})    & 0.10$^*$ (\textbf{0.87})     & 0.84 (0.16)     & 0.68 (0.31)    & 0.85 (0.15)     & 0.76 (0.23)    \\
Camou   & 0.63 (0.37)     & 0.49 (0.46)    & 0.82 (0.18)    & 0.53 (0.44)     & 0.98 (0.02)      & 0.74 (0.25)    & 0.99 (0.01)        & 0.78 (0.21)      \\
CAC & 0.08 (\textbf{0.92})    & 0.07 (\textbf{0.88})   & 0.17 (0.83)    & 0.19 (0.78)     & 0.58 (\textbf{0.42})      & 0.48 (\textbf{0.51})    & 0.66 (\textbf{0.34})     & 0.63 (\textbf{0.36})           \\ \hline
\end{tabular}
\caption{Average precision P@0.5 and drop rates of different attacks in restricted and free viewpoints. Each P@0.5 is averaged over two brightness levels. * indicates the white-box attacks. The best results are in bold.}
\label{tab:4}
\end{table*}

\section{More Experimental Results}
\subsection{Attacking the PascalVOC Dataset Models }

Tables \ref{tab:2} and \ref{tab:4} show the attack results for VGG16 and RseNet101 models trained on PascalVOC datasets, {\it i.e.}, FR-VGG16-07, FR-VGG16-0712, FR-RES101-07 and FR-RES101-0712. The model structures and training datasets of these models are different from the source model FR-Incv2-14.

\subsection{Qualitative Samples in Virtual Scene}
Figures \ref{fig:9} to \ref{fig:12} show several detection results of different models for the vehicles with camouflages generated by different methods under different viewing and lighting conditions. Figure \ref{fig:13} shows the results of simple textures under different viewpoints and brightness conditions. 

\subsection{Qualitative Samples in Physical Environment}
Figure \ref{fig:14} shows the results of vehicles with the CAC adversarial camouflage textures in the real world. Video of this experiment is available in the supplemental files.

\subsection{Generalization to Other Objects and Classes}
Figure \ref{fig:15} shows the results of other objects ({\it e.g.}, a barrel and a container) with the adversarial camouflage textures. Figure \ref{fig:16} shows vehicles with other target label ({\it e.g.}, stop sign) adversarial camouflage textures generated by the CAC attack. 

\renewcommand\thefigure{9}
\begin{figure*}[!htb]
\centering
\includegraphics[width=0.84\textwidth]{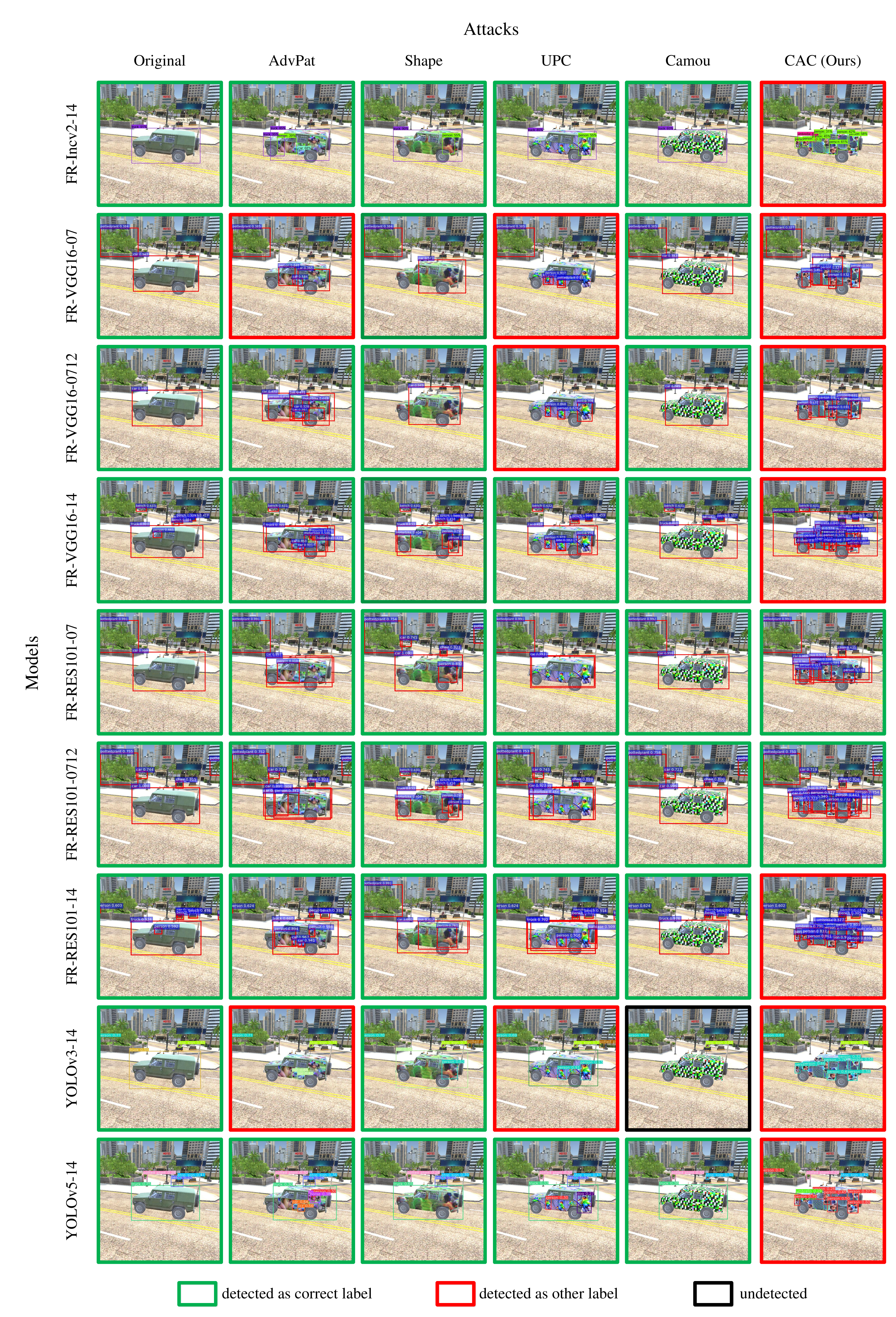} 
\caption{A sample of the 3D adversarial vehicles under the bright environment condition in the left side view. Zoom in for more details.}
\label{fig:9}
\end{figure*}

\renewcommand\thefigure{10}
\begin{figure*}[!htb]
\centering
\includegraphics[width=0.84\textwidth]{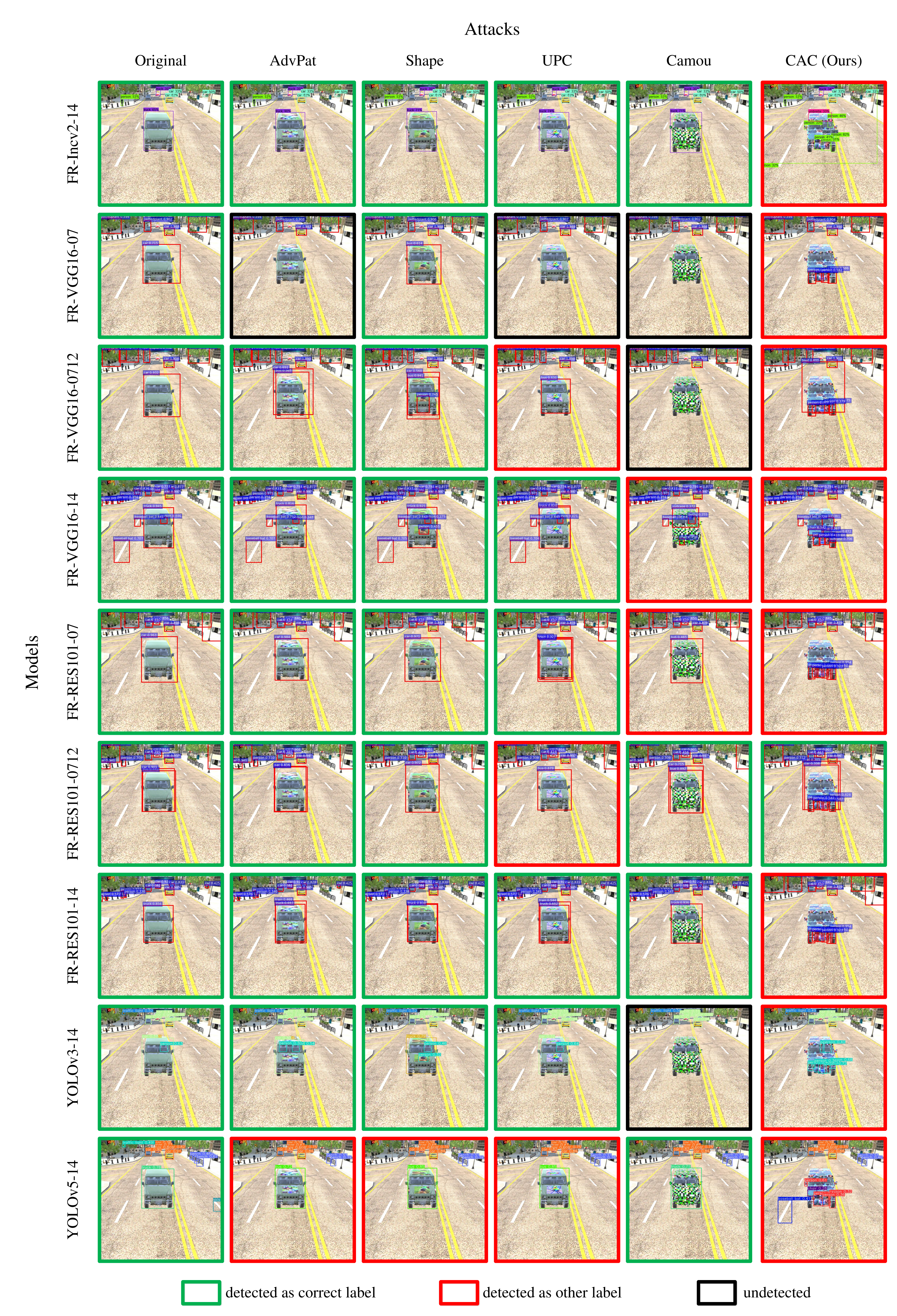} 
\caption{A sample of the 3D adversarial vehicles under the bright environment condition in the front view. Zoom in for more details.}
\label{fig:10}
\end{figure*}

\renewcommand\thefigure{11}
\begin{figure*}[!htb]
\centering
\includegraphics[width=0.84\textwidth]{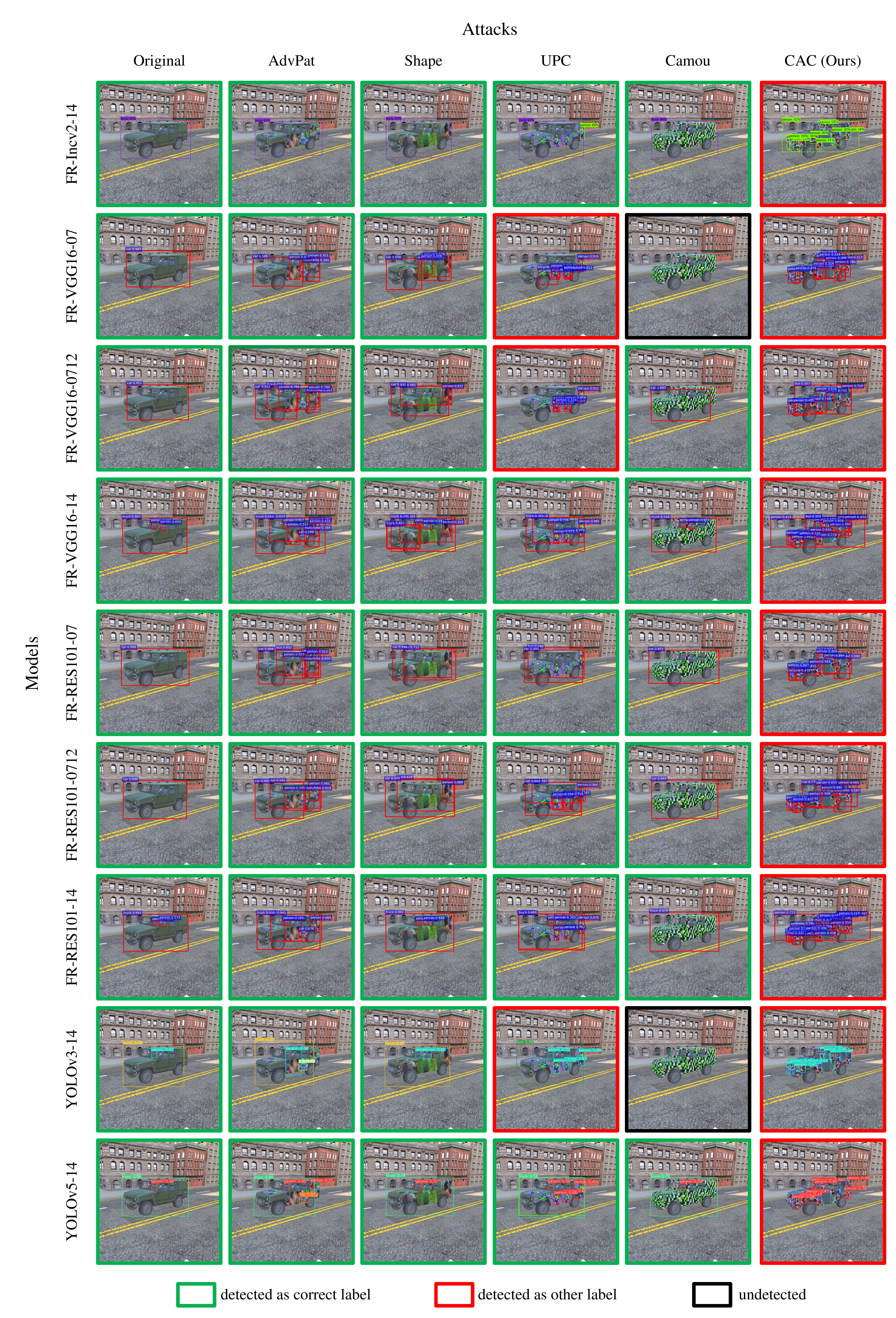} 
\caption{A sample of the 3D adversarial vehicles under the dark environment condition in the left view. Zoom in for more details.}
\label{fig:11}
\end{figure*}

\renewcommand\thefigure{12}
\begin{figure*}[!htb]
\centering
\includegraphics[width=0.84\textwidth]{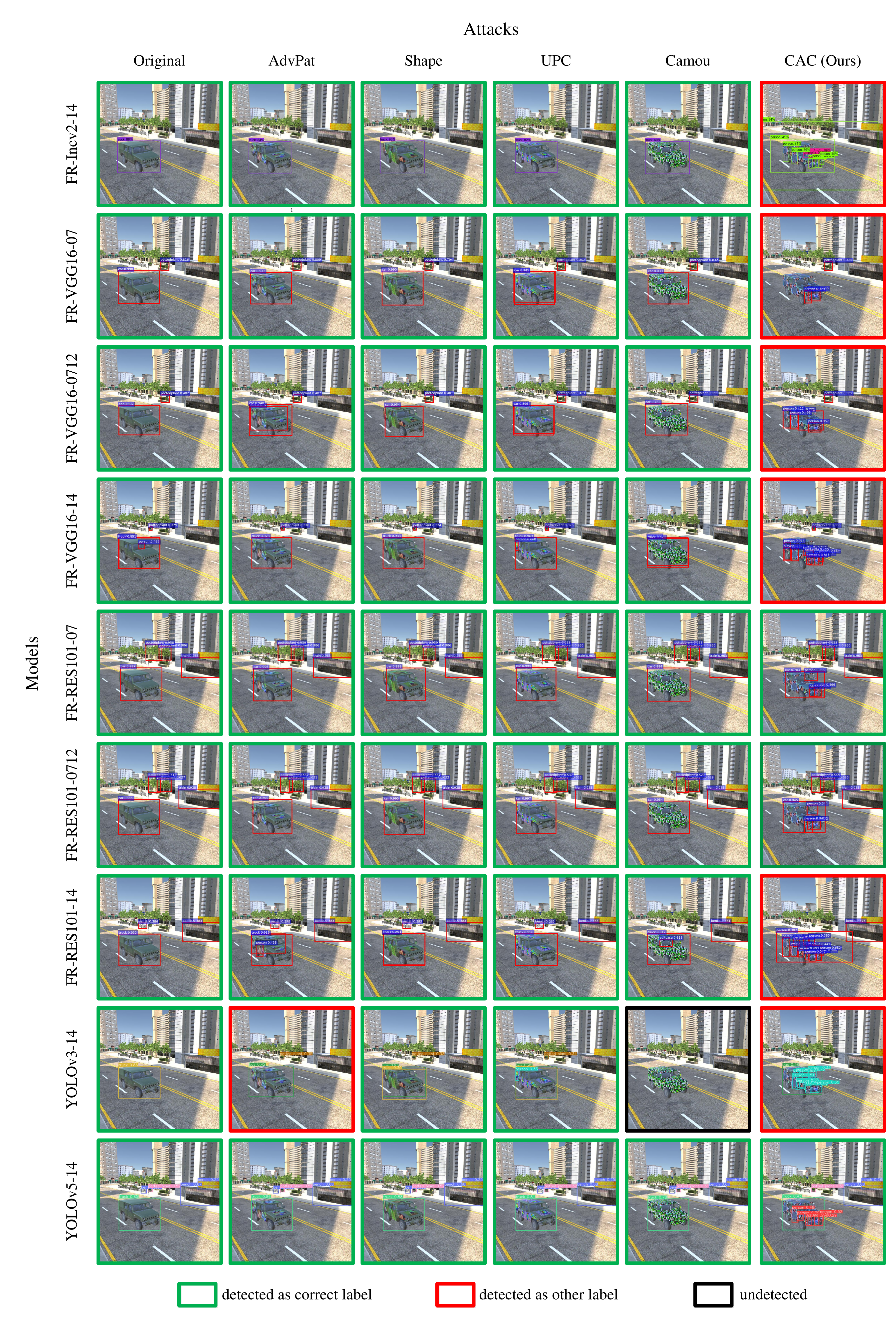} 
\caption{A sample of the 3D adversarial vehicles under the dark environment condition in the right front view. Zoom in for more details.}
\label{fig:12}
\end{figure*}

\renewcommand\thefigure{13}
\begin{figure*}[!htb]
\centering
\includegraphics[width=0.75\textwidth]{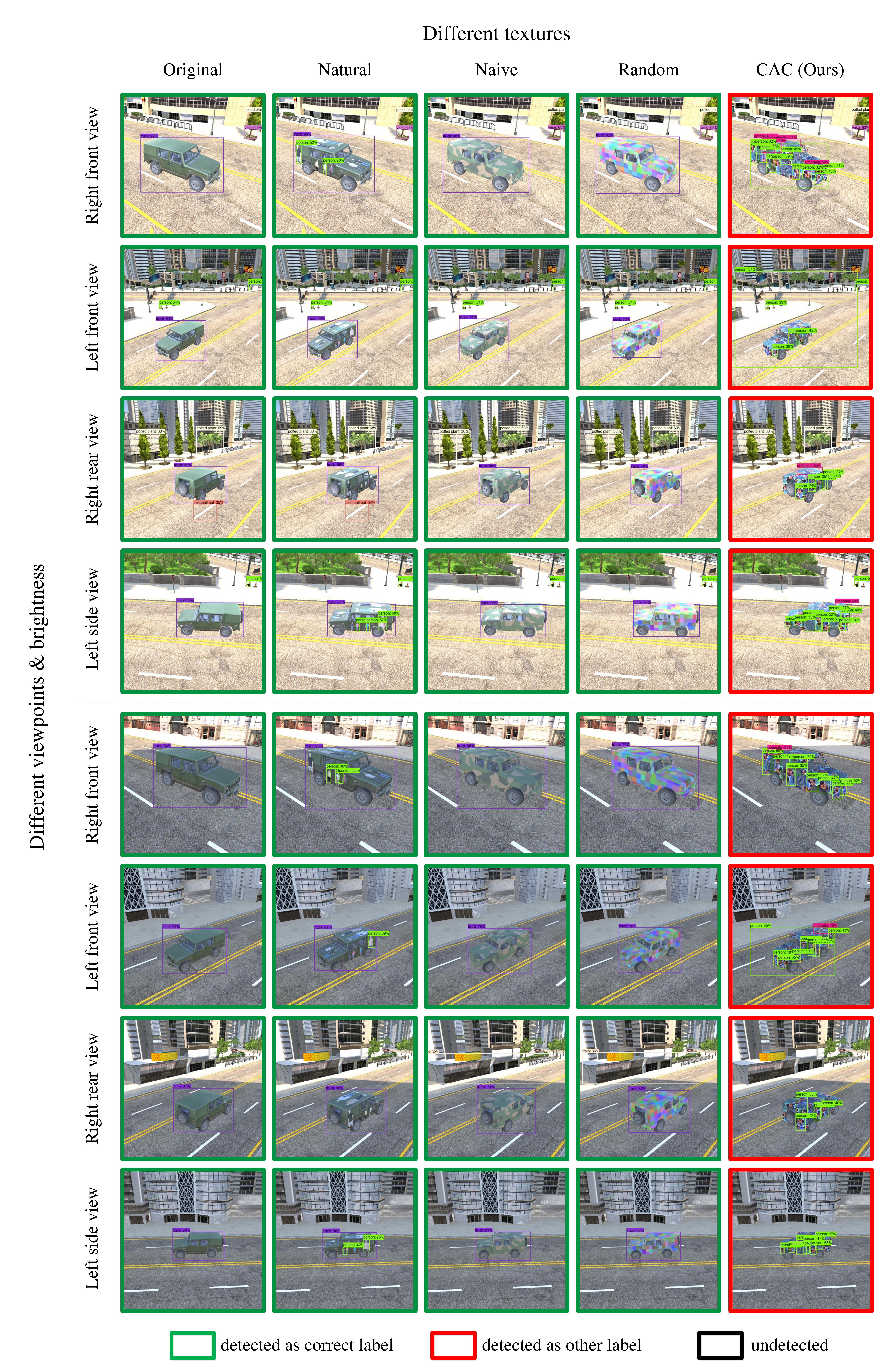} 
\caption{A sample of the 3D vehicles with simple textures and our CAC adversarial textures. Zoom in for more details.}
\label{fig:13}
\end{figure*}

\renewcommand\thefigure{14}
\begin{figure*}[!htb]
\centering
\includegraphics[width=0.75\textwidth]{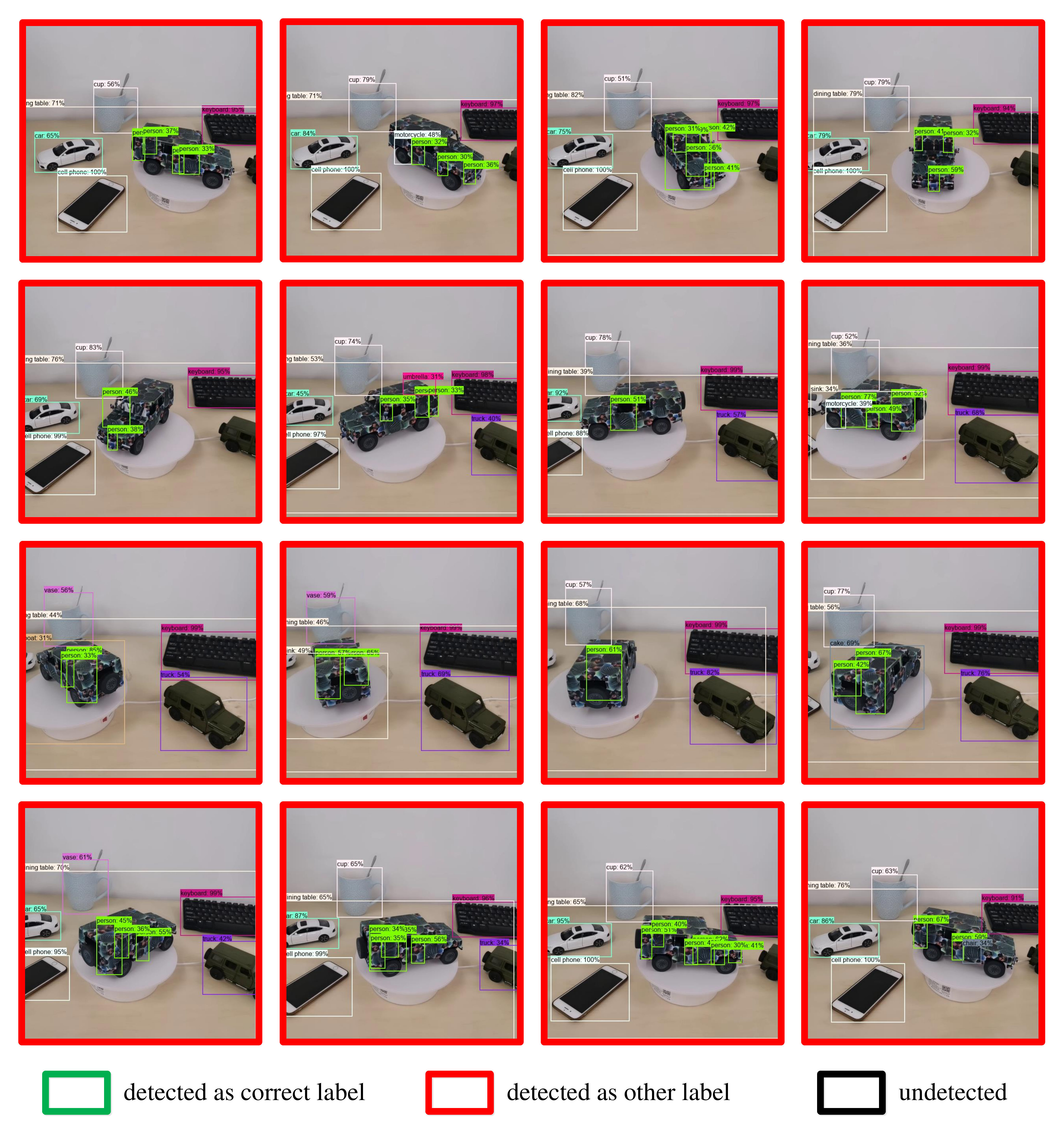} 
\caption{A sample of the adversarial vehicles in the real world. The results further demonstrate that our attack is invariant to different viewpoints. Zoom in for more details.}
\label{fig:14}
\end{figure*}

\renewcommand\thefigure{15}
\begin{figure*}[!htb]
\centering
\includegraphics[width=0.75\textwidth]{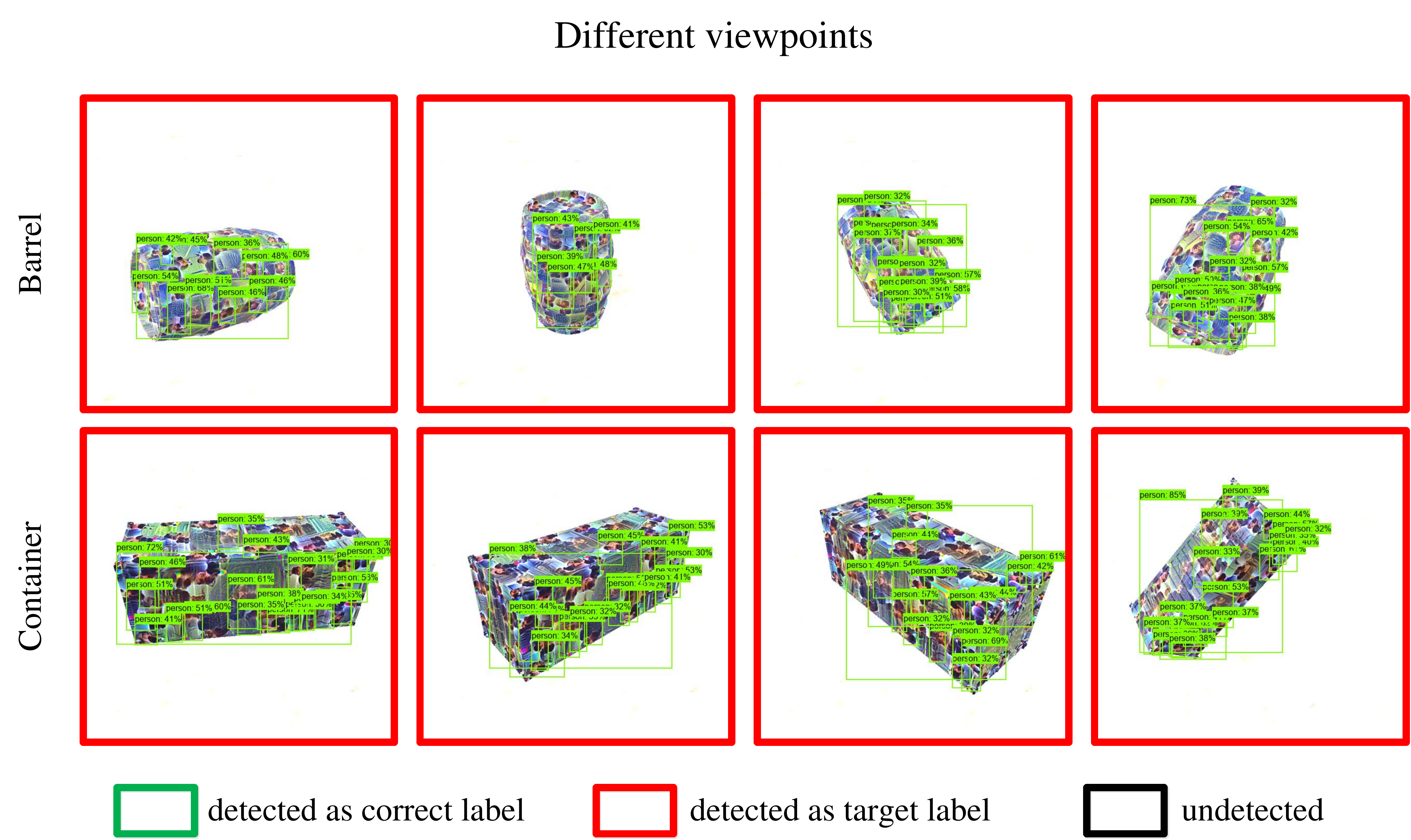} 
\caption{The results of attacking a barrel and a container. Zoom in for more details.}
\label{fig:15}
\end{figure*}

\renewcommand\thefigure{16}
\begin{figure*}[!htb]
\centering
\includegraphics[width=0.85\textwidth]{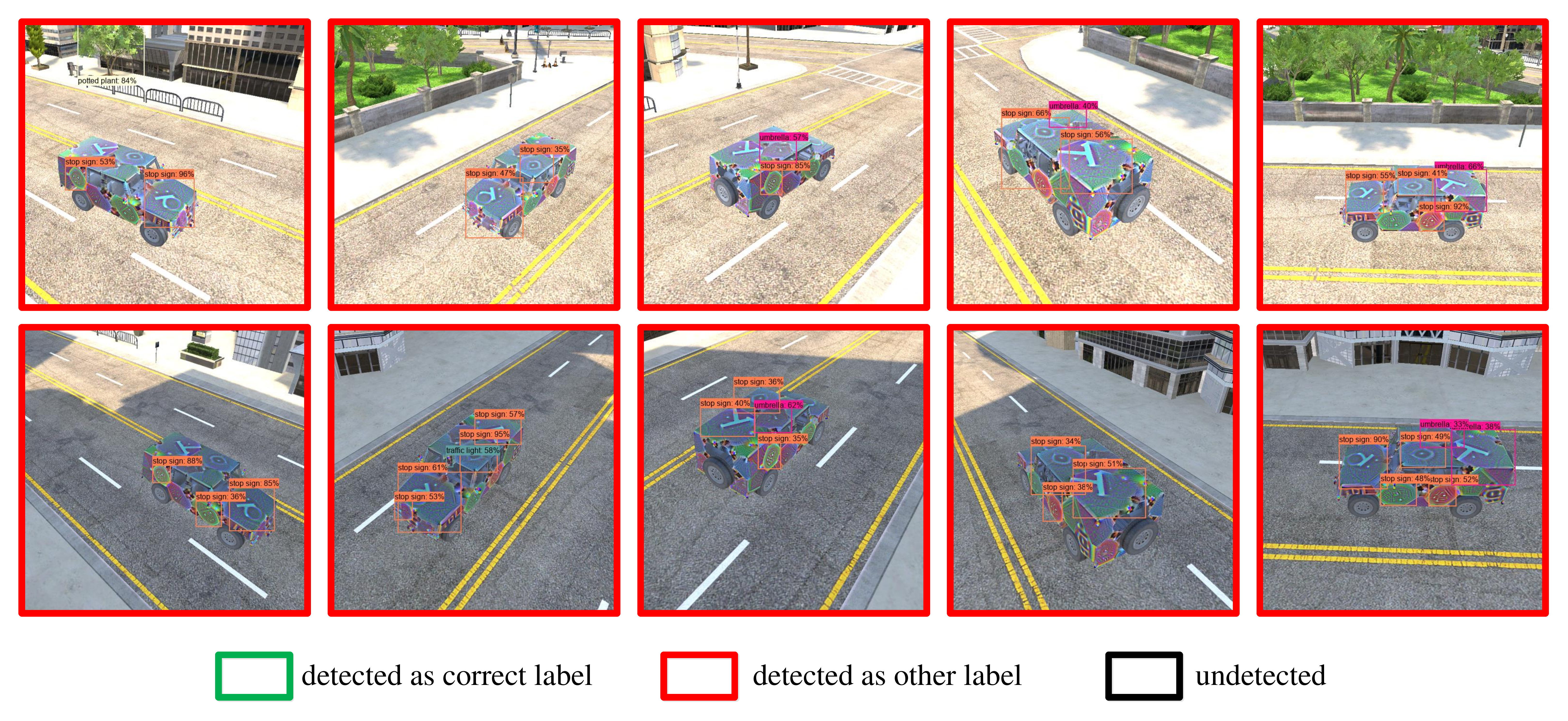} 
\caption{A sample of the 3D vehicles with other target label (stop sign) adversarial camouflages generated by CAC. Zoom in for more details.}
\label{fig:16}
\end{figure*}

\end{document}